\documentclass{article} % For LaTeX2e
\usepackage[final]{colm2026_conference}

\usepackage{microtype}
\usepackage{hyperref}
\usepackage{url}
\usepackage{booktabs}
\usepackage{marvosym}
\usepackage[most]{tcolorbox}
\usepackage{inconsolata}

\usepackage{graphicx}
\usepackage{enumitem}

\usepackage{wrapfig}

\usepackage{booktabs}   % \toprule \midrule \bottomrule
\usepackage{multirow}   % \multirow
\usepackage{array}      % better column control
\usepackage{colortbl}   % \cellcolor
\usepackage[table]{xcolor} % colors in tables (cellcolor)
\usepackage{amsmath} 
\usepackage{pifont}
\usepackage{xcolor}
\usepackage{booktabs}
\usepackage{array}
\usepackage{fancyvrb} % for Verbatim inside tables
\usepackage{fontawesome5}

\definecolor{checkgreen}{RGB}{0,140,0}
\definecolor{crossred}{RGB}{180,0,0}

\newcommand{\open}{\textcolor{checkgreen}{\ding{52}}}   % ✓ green
\newcommand{\closed}{\textcolor{crossred}{\ding{56}}}   % ✗ red
\usepackage{lineno}

\definecolor{darkblue}{rgb}{0, 0, 0.5}
\hypersetup{colorlinks=true, citecolor=darkblue, linkcolor=darkblue, urlcolor=darkblue}

\usepackage[most]{tcolorbox}
\tcbuselibrary{listings,breakable,skins}
\usepackage{listings}
\usepackage{xcolor}

\lstdefinelanguage{json}{
  basicstyle=\ttfamily\small,
  showstringspaces=false,
  breaklines=true,
  frame=none,
  columns=fullflexible,
  morestring=[b]",
  stringstyle=\color{teal!70!black},
  commentstyle=\color{gray},
  keywordstyle=\color{blue!60!black},
  literate=
   *{0}{{{\color{purple!70!black}0}}}{1}
    {1}{{{\color{purple!70!black}1}}}{1}
    {2}{{{\color{purple!70!black}2}}}{1}
    {3}{{{\color{purple!70!black}3}}}{1}
    {4}{{{\color{purple!70!black}4}}}{1}
    {5}{{{\color{purple!70!black}5}}}{1}
    {6}{{{\color{purple!70!black}6}}}{1}
    {7}{{{\color{purple!70!black}7}}}{1}
    {8}{{{\color{purple!70!black}8}}}{1}
    {9}{{{\color{purple!70!black}9}}}{1}
    {.}{{{\color{purple!70!black}.}}}{1}
    {-}{{{\color{purple!70!black}-}}}{1}
    {:}{{{\color{black}:}}}{1}
    {,}{{{\color{black},}}}{1}
    {\{}{{{\color{black}\{}}}{1}
    {\}}{{{\color{black}\}}}}{1}
    {[}{{{\color{black}[}}}{1}
    {]}{{{\color{black}]}}}{1}
}

\newtcolorbox{jsonbox}[1][]{
  breakable,
  colback=black!2,
  colframe=black!30,
  boxrule=0.6pt,
  arc=2pt,
  left=6pt,right=6pt,top=6pt,bottom=6pt,
  listing only,
  listing options={language=json},
  title=\textbf{Example CreditQA Question Record (JSON)},
  #1
}

\title{Credit Cards, Confusion, Computation, and Consequences:
What Can We Uncover About Language Model Reasoning?
 }

\author{Arnav Hiray\textsuperscript{\Letter}, Agam Shah\textsuperscript{\Letter}, Caleb Lu,
Meghaj Tarte, Harsit Mittal, \& Sudheer Chava\\
College of Computing \& Scheller College of Business\\
Georgia Institute of Technology, Atlanta, GA \\
\texttt{\Letter\;Equal First Authors: \{ahiray3, ashah482\}@gatech.edu}}

\begin{document}

\ifcolmsubmission
\linenumbers
\fi

\maketitle

\begin{abstract}
We introduce \textbf{\textsc{CreditCardQA}}\footnote{Our data and can be found on \href{https://hf.co/collections/gtfintechlab/creditqa}{Hugging Face} and \href{https://github.com/gtfintechlab/CreditQA}{Github}}, the first financial literacy benchmark for numerical reasoning derived from real credit card agreements. The dataset contains 1,800 questions, including first-person variants that reflect how consumers naturally ask about fees, interest, and payments. We evaluate a range of large language and reasoning models under Chain-of-Thought (CoT) and Program-of-Thought (PoT) prompting. Overall, PoT yields consistent performance gains, particularly for models with weaker baseline reasoning, and narrows gaps between open- and closed-source systems. Through error analysis, we show that failures arise less from arithmetic and more from misapplied financial rules, missed conditions, and misunderstandings of contractual terms. We further analyze question difficulty and find that comparisons, conditional logic, and monetary constraints are especially challenging. We also find that errors often arise in edge cases such as late-payment penalties or small-balance scenarios \textit{that are more likely to affect lower-income or financially vulnerable individuals}.
\end{abstract}

\section{Introduction}

\noindent \textcolor{red}{\textbf{\$1.2 trillion.}} This is the amount of credit card debt currently owed by consumers in the United States. 
%In roughly the time it takes to read this sentence, total household debt increases by about \$19,000.\footnote{Estimated from a \$24 billion quarterly increase in credit card debt reported by the Federal Reserve Bank of New York \url{https://www.newyorkfed.org/microeconomics/hhdc}.} 
Poor consumer borrowing behavior is a symptom of a far more widespread and systemic issue: persistently low levels of financial literacy. Survey-based evidence from the World Economic Forum and the Personal Finance Index (P-Fin Index) indicates that financial literacy in the United States has remained near 50\% for eight consecutive years, with no sustained improvement over time \citep{WEF2024FinancialLiteracy,GFLEC_PFinIndex}. Fewer than one-third of U.S. adults correctly answer basic questions on interest compounding, inflation, and risk diversification. While this issue spans all ages, financial literacy is lowest among younger generations on average, indicating persistent gaps early in life.

The causal relationship between financial literacy and financial health is well established. A large body of research demonstrates that low financial literacy is consistently associated with excessive debt \citep{NBERw17103}, higher rates of bankruptcy \citep{brown2016financial}, and inadequate saving for retirement or emergencies \citep{NBERw17108}. More concerning, unequal financial outcomes persist across populations. Evidence from consumer credit markets shows that women tend to start out with lower credit limits than men when they first enter the credit card market, in part because they sort into relatively inferior credit card products \citep{Ganduri2023GenderSorting}. Additionally, financial literacy and access to effective financial education remain lower among vulnerable and historically underserved populations \citep{GFLEC_PFinIndex}. Together, these disparities amplify the consequences of low financial literacy, making gaps in financial understanding particularly costly for already disadvantaged groups.

% Low-literacy cardholders are disproportionately caught in debt spirals, where missed payments lead to mounting interest and a cycle of borrowing they cannot escape \citep{Giannikos2025}. 

Financial education initiatives have become the primary approach to addressing gaps in financial literacy. However, while empirical evidence shows that state-of-the-art financial education methods can improve financial knowledge and behavior \citep{NBERw27057}, existing programs are often costly to implement, improve financial behavior by a small margin, and yield highly uneven outcomes across individuals, limiting the effectiveness of the one-size-fits-all approaches commonly used in these studies \citep{McKenzie2022FinancialLiteracy}. However, the most effective financial education delivers guidance \textbf{at the moment} individuals are making financial decisions as \emph{``teachable moments"}, \textbf{tailored to their specific circumstances} rather than relying on generic instruction \citep{WorldBank2017FinancialEducation}.

Recent advances in large language models (LLMs) closely align with the characteristics required for effective financial education. Evidence from large-scale usage data shows that a majority of interactions with ChatGPT involve \emph{Practical Guidance} or \emph{Seeking Information}, suggesting that users frequently engage with these systems to ask concrete questions and obtain assistance at the moment of need \citep{NBERw34255}. Importantly, adoption of these tools spans a broad and increasingly diverse user base, with substantially reduced gender and socioeconomic skew over time \citep{NBERw34255}. Beyond general use, survey and reporting evidence indicates that consumers are increasingly relying on AI systems for financial decision-making, with more than half of adults and roughly two-thirds of Gen Z and Millennials using these tools for personal finance tasks \citep{Hosking2025AIAsFinancialAdviser, Sola2024AIForPersonalFinance}.

Finance is widely considered a high-stakes domain, demanding higher accuracy in settings where errors can have significant consequences. Consequently, financial numerical reasoning has received substantial attention, with a growing body of work examining LLM performance on finance-related tasks \citep{tang-etal-2025-financereasoning, chen-etal-2021-finqa, zhao-etal-2024-knowledgefmath, zhao-etal-2024-tapera, shah2023zeroheroyetbenchmarking, peng2025multifinbenmultilingualmultimodaldifficultyaware, zhao-etal-2024-docmath, zhu-etal-2021-tat, krumdick-etal-2024-bizbench, hiray-etal-2024-cocohd, chen-etal-2022-convfinqa, shah-etal-2024-numerical}. Notably, none of these benchmarks are designed to assess financial literacy; instead, they focus directly or indirectly on stock market related reasoning, including investment analysis, firm performance, and professional finance tasks.

% This line of research has revealed persistent bottlenecks, including limited business and financial understanding, trade-offs between cost and performance as model scale increases, and recurring numerical calculation errors. 

We believe that personal finance represents an equally if not more consequential high-stakes financial setting, affecting a substantially larger population and directly shaping financial well-being and quality of life. Despite growing interest in financial reasoning, there remains no prior systematic evaluation of whether LLMs can support the reasoning required for real-world financial literacy tasks. As LLMs are increasingly used for personal financial decision-making, assessing their reliability in this setting is both important and urgent.

To address this gap, we introduce \textbf{CreditCardQA }, a financial numerical reasoning dataset consisting of 1,800 question–answer pairs, which, to the best of our knowledge, is the first designed to evaluate LLMs' ability to perform numerical reasoning in real-world financial literacy scenarios. We construct questions directly from credit card agreements, which specify the terms and conditions governing card usage, including interest accrual, payment rules, fees, and penalties. We ask the following research questions: 

\begin{itemize}
\itemsep 0em
    \item \textbf{RQ1:} How well do large language models (LLMs) and large reasoning models (LRMs) perform on real-world financial literacy numerical reasoning questions?
    \item \textbf{RQ2:} What types of numerical and reasoning errors do these models make when answering questions grounded in credit card agreements?
    \item \textbf{RQ3:} How can the difficulty of financial literacy questions be systematically assessed, and how does model performance vary with question difficulty and reasoning complexity?
\end{itemize}

\begin{figure}[t]
    \centering
    \includegraphics[width=\linewidth]{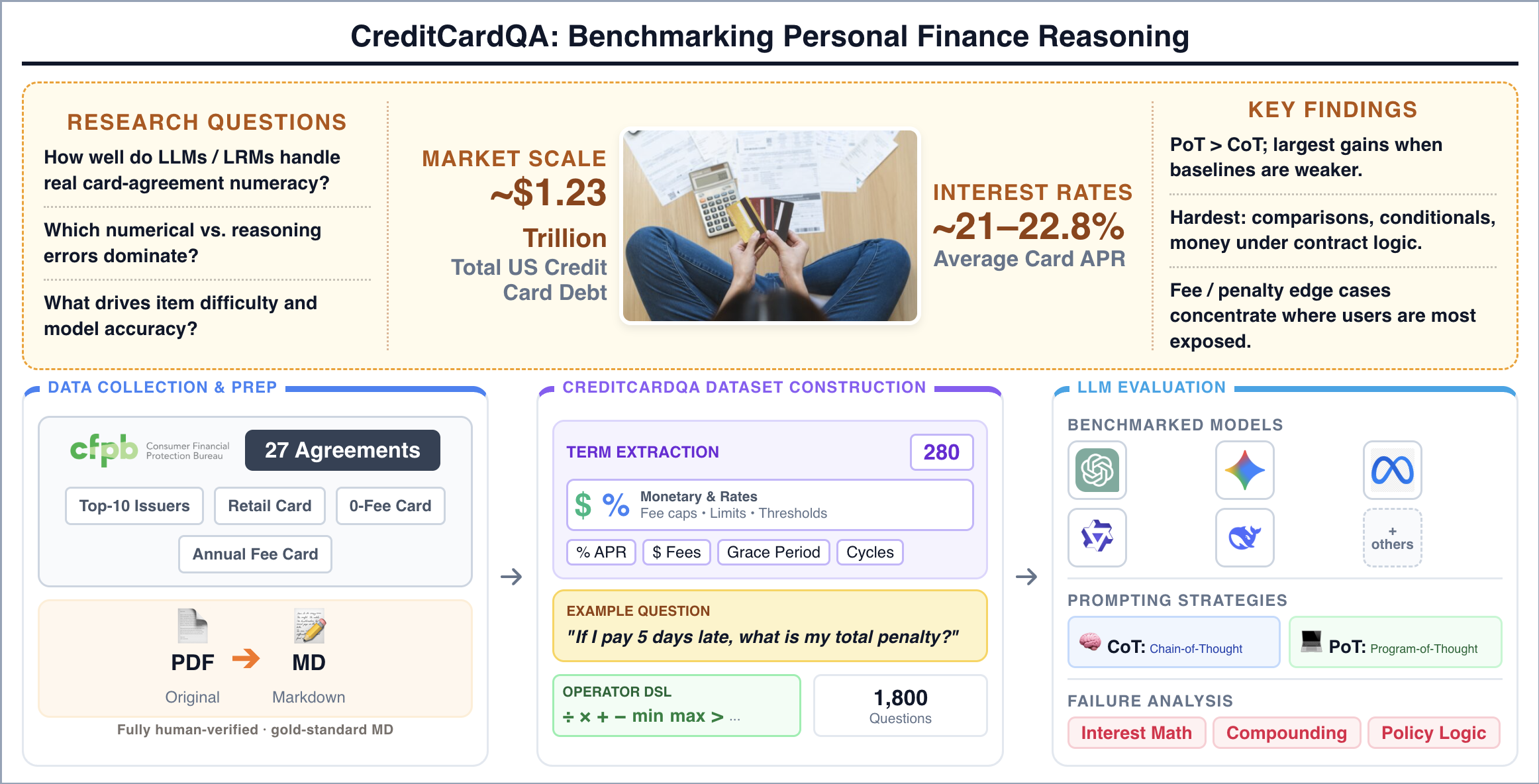}
    \caption{\textsc{CreditCardQA } construction pipeline: CFPB credit card agreements $\rightarrow$ 280 extracted terms $\rightarrow$ 1,800 generated questions $\rightarrow$ LLM evaluation with CoT/PoT prompting.}
    \label{fig:fig1}
\end{figure}

\section{\textsc{CreditQA} Construction}
To facilitate our analysis, we construct \textsc{FinLitQA}, a dataset of numerical reasoning questions grounded in real-world credit card agreements (Section~2.1), and detail the question generation and annotation process in Section~2.2.

\subsection{Credit Card Agreements}
A credit card agreement is a legally binding document that specifies the terms and conditions governing a credit card account, including interest rates, fees, payment obligations, penalties, and consumer rights. These agreements are often many pages long and written at a reading level well beyond that of the average U.S. consumer, making them difficult to interpret in practice \citep{Schorn2016CreditCard}.

\begin{quote}\small
\emph{``The Minimum Payment Due is 2.00\% of Your total New Balance, or \$15.00, whichever is greater, plus any amount past due and any amount by which You have exceeded Your applicable credit limit. If Your total New Balance is less than \$15.00, then Your Minimum Payment Due is the amount of the total New Balance.''}

\hfill {\footnotesize -- Excerpt from a credit card agreement describing minimum payment terms.}
\end{quote}

As illustrated in the excerpt above, credit card agreements contain highly involved terms and conditions, incorporating multiple financial concepts, numerical values, and mathematical operators (e.g., comparisons and arithmetic). While these documents are widely known to be difficult for humans to understand, we show throughout this work that language models too exhibit weakness. Credit card agreements serve as the primary (and often sole) source of information available to cardholders for learning how to use their cards and understand the associated terms and costs. For this reason, they also serve as the sole source documents from which we derive all numerical reasoning questions in our dataset.

\textbf{Selecting Credit Card Agreements} 
We collected 27 credit card agreements from the \citep{CFPB2024Agreements} (CFPB) database, covering major issuers such as American Express and Discover. To introduce more diversity beyond the U.S. market, we additionally include Barclays, a major European card issuer that serves a substantial portion of European consumers. Agreements were selected from Q4 2024 to account for changes in terms and conditions, and issuer selection was guided by market share using external rankings based on outstanding credit card balances \citep{WalletHub2024MarketShare}. The issuers included in our dataset collectively account for over 80\% of total outstanding balances \footnote{Outstanding balance denotes the total unpaid credit card debt held by an issuer across all active card accounts at a given time. Source: \href{https://wallethub.com/edu/cc/market-share-by-credit-card-issuer/25530}{Credit Card Market Share by Issuer}
}, ensuring broad coverage of real-world credit card usage rather than random sampling. For each issuer, we included agreements from three product categories: premium/elite (comes with annual fees), everyday retail, and subprime cards (with no annual fees) to capture heterogeneity in card design and usage.

In the CFPB database, credit card agreements are provided in PDF format and often include images, tables, and text. However, to make them accessible for LLMs, we converted all agreements into Markdown format using Marker and manually verified their quality. Additionally, to ensure fairness, the question generation process relied solely on these Markdown versions so that no information outside the models’ accessible context was used.

\subsection{\textsc{CreditQA} Question Creation}
The dataset questions are created by a team of annotators with expertise in finance and machine learning, including PhD holders, researchers working in machine learning for finance, contributors holding a Master’s degree in Quantitative Finance, individuals with CFA-level financial knowledge, and contributors with experience in hedge funds, investment banking, and asset management, under the guidance of a Chair Professor of Finance. All stages of dataset development were overseen by the core authors, and the process spanned approximately ten months. The question creation process was divided into five key steps as detailed in Appendix~\ref{ap:question_creation}.

\begin{wraptable}{r}{0.5\textwidth}
\vspace{-.4cm}
\centering
\scriptsize
\renewcommand{\arraystretch}{.8}
\begin{tabular}{p{0.7\linewidth} r}
\toprule
\multicolumn{2}{c}{\textbf{Credit Card Agreements}} \\
\midrule
\# of Financial Numeric Terms & 280 \\
\# of Words (Med / Avg) & 9115 / 8830 \\
\# of Percentages (Med / Avg) & 20 / 22.6 \\
\# of Dollar Amounts (Med / Avg) & 15.5 / 19.25 \\
\# of Bullet Points (Med / Avg) & 55.5 / 68.6 \\
Flesch--Kincaid Readability Score (Min / Avg) & 9.2 / 11.3 \\
\midrule
\multicolumn{2}{c}{\textbf{\textsc{CreditQA} Dataset}} \\
\midrule
Number of (Third/First) Person Questions & 1308/492 \\
Question Length (Med / Avg) & 33 / 36.7 \\
\# of Math Operations Required (Med / Avg) & 3 / 4.4 \\
\# of Document Inputs (Max/Avg) & 6/1.0 \\
\# of User Inputs (Max/Avg) & 10/2.0 \\
Dev/Test Question Split & 800/1000 \\
\bottomrule
\end{tabular}
\caption{Statistics of Credit Card Agreements and \textsc{CreditQA} Dataset.}
\label{tab:cca_stats}
\vspace{-.4cm}
\end{wraptable}

\usetikzlibrary{calc}

\tcbset{
  inputbox/.style={
    colback=gray!5!white,
    colframe=black!35,
    boxrule=0.4pt,
    arc=2pt,
    left=3pt, right=3pt, top=3pt, bottom=3pt,
    fontupper=\scriptsize
  },
  llmbox/.style={
    colback=blue!4!white,
    colframe=blue!45!black,
    boxrule=0.4pt,
    arc=2pt,
    left=3pt, right=3pt, top=3pt, bottom=3pt,
    fontupper=\scriptsize
  }
}

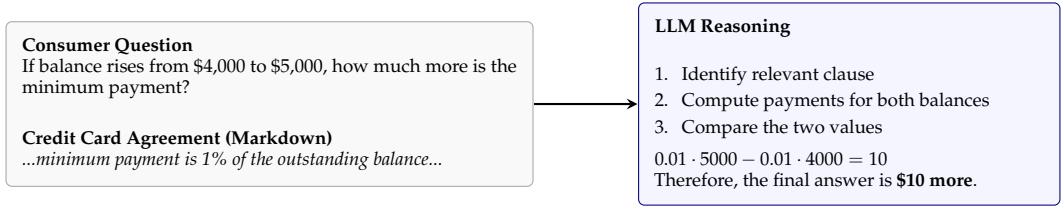
\begin{figure}[t]
\centering
\begin{tikzpicture}[>=stealth, every node/.style={inner sep=0pt}]

\node[anchor=west] (inp) at (0,0) {
\begin{tcolorbox}[inputbox, width=0.5\columnwidth]
\textbf{Consumer Question}\\
If balance rises from \$4{,}000 to \$5{,}000, how much more is the minimum payment?\\[3pt]

\textbf{Credit Card Agreement (Markdown)}\\
\textit{...minimum payment is 1\% of the outstanding balance...}
\end{tcolorbox}
};

\node[anchor=east] (llm) at (\columnwidth,0) {
\begin{tcolorbox}[llmbox, width=0.4\columnwidth]
\textbf{LLM Reasoning}\\[-2pt]
\begin{enumerate}[leftmargin=*, itemsep=0.0em]
    \item Identify relevant clause
    \item Compute payments for both balances
    \item Compare the two values
\end{enumerate}

{\scriptsize $0.01 \cdot 5000 - 0.01 \cdot 4000 = 10$}

Therefore, the final answer is \textbf{\$10 more}.
\end{tcolorbox}
};

\draw[->, thick] (inp.east) -- (llm.west);

\end{tikzpicture}
\caption{Simplified process for answering CreditCardQA questions from credit card agreements.}
\label{fig:finlitqa_pipeline}
\end{figure}
\subsection{Data Statistics and Dataset Release}
\textbf{Dataset Statistics} Table~\ref{tab:cca_stats} summarizes key statistics of the dataset and the underlying credit card agreements. One notable observation is the low Flesch–Kincaid readability scores of these agreements, which correspond to reading levels typically associated with college graduates or professionals \citep{wiki:flesch}. Furthermore, in comparison to other state-of-the-art financial QA datasets such as \citet{tang-etal-2025-financereasoning}, our average number of mathematical operators per question is slightly higher, indicating comparable reasoning complexity from a mathematical perspective. Furthermore, our dataset contains 821 unique operator sequences and 602 unique input–output combinations, highlighting the diversity of reasoning patterns and question types. Appendix~\ref{ap:question_format} shows an example question with complete annotations.

Importantly, while prior financial reasoning benchmarks include concepts such as interest rates or APR, they are typically presented in standardized, formula-driven settings. In contrast, credit card agreements introduce substantially greater complexity, defining multiple APR types (e.g., purchase, cash advance, and penalty APR), each governed by distinct conditions, alongside constructs such as billing cycles, grace periods, minimum payment rules, and promotional terms. Consequently, these questions require not only numerical computation but also precise interpretation of agreement-specific terminology and conditional rules—particularly logical structures such as \emph{if} statements, comparisons (e.g., greater than), and operations like max/min and conjunctions (and/or), which are largely underexplored in prior financial reasoning benchmarks and constitute a central focus of our dataset.

\textbf{Dataset Release} We divide the dataset into a \emph{development} set and a \emph{test} set. The test set consists of 1,000 questions that will be maintained on an online leaderboard upon acceptance. The remaining 800 questions, along with their fully labeled annotations, are publicly available for model development. Compared to prior work, we adopt a more lenient split that allocates a larger portion of questions to development. While maintaining a held-out test set remains essential for meaningful comparison, this choice ensures sufficient data for analysis and facilitates future dataset extensions using this methodology. The use of held-out test sets is also consistent with prior work in numerical reasoning and financial QA \citep{chen-etal-2021-finqa, chen-etal-2022-convfinqa, zhu-etal-2021-tat, islam2023financebench, zhao-etal-2024-knowledgefmath, zhao-etal-2024-docmath}.

\section{Evaluation}
We evaluate 11 open- and closed-source models spanning both Large Language Models (LLMs) and Large Reasoning Models (LRMs).

\textbf{Models.}
\textbf{Open-weight:} DeepSeek-R1 \citep{deepseekai2025deepseekr1incentivizingreasoningcapability}, Qwen-QwQ \citep{qwq32b}, GPT-OSS-20B, GPT-OSS-120B \citep{agarwal2025gpt}, Kimi-K2-Thinking \citep{team2025kimi}, Llama-3.3-70B \citep{grattafiori2024llama3herdmodels}, Mistral-Small-24B \citep{mistralai2025mistral_small_3}, and GLM-4.6 \citep{zeng2025glm}. 
\textbf{Closed-source:} GPT-5, GPT-5-mini \citep{openai2025gpt5}, and Gemini 3.0 Pro \citep{geminiteam2024geminifamilyhighlycapable}.

\textbf{Prompting.}
Following \citet{tang-etal-2025-financereasoning,zhao-etal-2024-knowledgefmath}, we evaluate Chain-of-Thought (CoT) \citep{wei2022chain} and Program-of-Thought (PoT) \citep{chen2023program}. Full details are in Appendix~\ref{ap:experiment-settings}.

\textbf{Evaluation.}
We adopt a similar answer-extraction pipeline as \citet{zhao-etal-2024-knowledgefmath}. For categorical outputs, we additionally perform manual verification (Appendix~\ref{ap:manual_annotation}). Accuracy is reported at relative error tolerances of 0.2\% and 5\%.
\section{Main Results (RQ1)}
\newcommand{\topperf}[1]{\cellcolor{green!15}{#1}}
\newcommand{\midperf}[1]{\cellcolor{yellow!15}{#1}}
\newcommand{\lowperf}[1]{\cellcolor{red!12}{#1}}

\begin{table*}[ht]
\centering
\footnotesize
\setlength{\tabcolsep}{5pt}
\renewcommand{\arraystretch}{1.05}

\begin{tabular}{@{}l c c cc cc c@{}}
\toprule
\textbf{Model} &
\textbf{Size} &
\textbf{Open Weight?} &
\multicolumn{2}{c}{\textbf{CoT Acc. (\%)}} &
\multicolumn{2}{c}{\textbf{PoT Acc. (\%)}} &
\textbf{$\Delta_{5\%}$ (PoT $-$ CoT)} \\
\cmidrule(lr){4-5}\cmidrule(lr){6-7}
& & &
$\pm$0.2\% &
$\pm$5\% &
$\pm$0.2\% &
$\pm$5\% &
\\
\midrule

\multicolumn{8}{@{}l}{\textit{Large Reasoning Models (LRMs)}}\\
Gemini 3.0 Pro & -- & \closed & \textbf{70.5} & \textbf{77.6} & 71.5 & \underline{79.1} & +1.5 \\
GPT-5 & -- & \closed & \underline{70.4} & \underline{76.7} & \textbf{72.4} & 79.0 & +2.3 \\
GPT-5-mini & -- & \closed & 70.0 & 75.8 & 68.9 & 76.7 & +0.9 \\
DeepSeek-R1 & 671B & \open & 63.3 & 70.9 & 69.8 & 76.6 & +5.7 \\
Qwen-QwQ & 32B & \open & 63.6 & 70.3 & 71.1 & 76.4 & +6.1 \\
GPT-OSS & 120B & \open & \text{70.0} & 76.9 & \underline{72.0} & \textbf{79.2} & +2.3 \\
GPT-OSS & 20B & \open & 66.8 & 73.7 & 67.6 & 74.9 & +1.2 \\
Kimi-K2-Thinking & 1T & \open & 64.1 & 71.9 & 68.4 & 76.4 & +4.5 \\
\midrule

\multicolumn{8}{@{}l}{\textit{Large Language Models (LLMs)}}\\
Llama 3.3 & 70B & \open & 62.0 & 69.6 & 70.9 & 75.8 & +6.2 \\
Mistral-Small & 24B & \open & 61.1 & 69.3 & 66.5 & 71.6 & +2.3 \\
GLM-4.6 & 357B & \open & 65.0 & 73.1 & 68.2 & 76.1 & +3.0 \\
\bottomrule
\end{tabular}

\caption{ \small
Main results on the \textsc{CreditQA} dataset.
We report \textbf{accuracy (\%)} under two numerical tolerances (±0.2\% and ±5\%) for Chain-of-Thought (CoT) and Program-of-Thought (PoT) prompting.
$\Delta_{5\%}$ (PoT$-$CoT) denotes the absolute accuracy improvement from CoT to PoT computed at the ±5\% tolerance.
Model rankings are highly consistent across tolerances within the same prompting strategy, but exhibit larger shifts between CoT and PoT, highlighting the impact of prompting choice on relative performance.
\textbf{Open Weight?} uses \open\ for open-weight and \closed\ for closed/proprietary models.
}

\label{tab:main_results}
\end{table*}

The performance of all models given prompting methods and error thresholds can be seen in Table~\ref{tab:main_results}. Overall, \textbf{GPT-OSS~120B, Gemini~3.0~Pro, and GPT-5} emerge as the top three models across all evaluated settings. Notably, the open-weight \textbf{GPT-OSS~120B} not only competes with but slightly outperforms leading closed-source models under Program-of-Thought, highlighting that strong open-weight systems can achieve state-of-the-art performance.  

\textbf{CoT vs.\ PoT.}
Across all  models, \emph{Program-of-Thought (PoT) prompting consistently outperforms Chain-of-Thought (CoT)}. Every model desmonstrates improvement from CoT to PoT, with gains ranging from 0.9\% to 6.2\% at the $\pm 5\%$ tolerance.

\textbf{Model class and sourcing effects.}
Closed-source models rank near the top across prompting strategies and error thresholds, with Gemini~3.0~Pro and GPT-5 achieving accuracy close to 80\% at the $\pm 5\%$ tolerance. Notably, open-weight GPT-OSS~120B performs comparably to these closed-source models and slightly surpasses them under PoT. While most open-weight models lag behind their closed-source counterparts under CoT, this gap narrows under PoT.

\textbf{LLMs vs.\ LRMs.}
The performance gap between large reasoning models (LRMs) and large language models (LLMs) depends strongly on the prompting strategy. Under CoT, LRMs outperform LLMs by approximately 4–5 percentage points at the $\pm 5\%$ tolerance, whereas \emph{under PoT this gap shrinks to roughly 2–3 percentage points}. 

\textbf{Who benefits most from PoT?}
PoT yields the largest gains for \emph{models with weaker baseline CoT performance}, including Llama~3.3, Qwen-QwQ, and DeepSeek-R1, while models with stronger CoT baselines (e.g., GPT-5-mini and GPT-OSS~20B) exhibit smaller improvements.

\textbf{Model size and family consistency.}
Larger models generally achieve higher accuracy, a pattern that holds within model families such as GPT-5 versus GPT-5-mini and GPT-OSS~120B versus GPT-OSS~20B.However, \textit{PoT gains are not monotonic in model size}: for example, the 32B Qwen-QwQ benefits more from PoT than the substantially larger GPT-OSS~120B, while the 671B-parameter DeepSeek-R1 benefits comparable to Qwen-QwQ.

\textbf{Retrieval Augmented Generation.} We additionally evaluate a simple retrieval-augmented generation (RAG) setting. Implementation details and results are provided in Appendix~\ref{ap:rag}.

\section{Error Analysis (RQ2)}

Beyond accuracy, we conduct a fine-grained error analysis of recurring failure modes in incorrect model responses, providing insight into how and why models fail on financial literacy questions. We focus on five non-mutually exclusive error types and examine both their prevalence and underlying causes. Specifically, we manually analyze 50 incorrect responses produced by GPT-OSS-120B, the strongest open-weight model in our evaluation. This analysis is intended as a focused audit of an accessible model that can support fine-tuning and follow-up interventions, rather than as a characterization of all 11 evaluated models, particularly closed frontier systems. Table~\ref{tab:error_class_method_license} reports the resulting distribution of error types. These qualitative trends are also consistent with our full-dataset regression analysis of GPT-OSS-120B, which identifies comparisons, branching, and monetary constraints as major sources of difficulty. Appendix~\ref{ap:glossary} defines the financial terms relevant to the analysis. We note the following:

\textbf{Formula Application Errors (Concept Substitution):} This is the most prevalent error type, occurring when the model identifies the correct financial concept but applies an incorrect formula or contractual rule. Common examples include using simple interest where compounding is required, substituting Purchase APR terms for Cash Advance APR terms, or misapplying Penalty APR criteria. Models also sometimes incorrectly treat fixed fees as interest-bearing balances. These errors suggest that while relevant agreement language is often retrieved correctly, models lack the domain-specific financial understanding.

\begin{table}

\centering
\footnotesize
\begin{center}
\begin{tabular}{l c c}
\toprule
\textbf{Error Type} & \textbf{\%} & \textbf{Count (/50)} \\
\midrule
Formula Application (Substitution) 
& \cellcolor{red!20} 70\% & 35 \\
Missing Condition Application 
& \cellcolor{orange!20} 54\% & 27 \\
Additional Unnecessary Steps 
& \cellcolor{yellow!20} 18\% & 9 \\
Numerical Calculation Error 
& \cellcolor{green!20} 14\% & 7 \\
Misunderstanding of Problem 
& \cellcolor{green!15} 8\% & 4 \\
\bottomrule
\end{tabular}
\end{center}
\caption{ \footnotesize
Distribution of error types (non-mutually exclusive) across 50 sampled questions.
\% denotes percentage indicating the fraction of questions exhibiting each error type; counts denote the number of affected questions.
Color 
reflects prevalence.
}
\label{tab:error_class_method_license}
\end{table}

\textbf{Missing Conditions:} These errors indicate that the model fails to identify or apply required conditional logic, such as minimum fees, state-specific caps, timing rules, or eligibility thresholds. As a result, the model performs a seemingly reasonable computation while ignoring an “if/only when” condition that determines whether the rule applies. More than half of the questions that have such an error indicate that correctly identifying and applying the terms and conditions specified in credit card agreements remains a challenge.

\textbf{Additional Unnecessary Steps:}
    Models often introduce superfluous assumptions or intermediate steps that are not required by the problem, resulting in overly complicated reasoning chains and incorrect answers. For example, a model may compute interest over a period even when the question only requires adding a fixed fee, or apply a Penalty APR without the triggering conditions being met. Although less common, this error may reflect inference-time overthinking induced by test-time scaling \citep{muennighoff2025s1}, where longer reasoning chains introduce unnecessary conditions or computations.

\textbf{Numerical Calculation Error:} In the majority of cases, the model correctly identified the relevant variables and retrieved the appropriate numerical values for each term. Instead, the numerical calculation errors we observed were primarily arithmetic or calendar-related mistakes, such as miscounting the number of days between two dates or assuming an incorrect number of days in a particular month (30 versus 31). The low prevalence of this error is encouraging, as it suggests that once the correct financial rules and inputs are identified, the model is generally capable of executing the required computations accurately.

\textbf{Misunderstanding of Problem:} This was the least frequent error type observed. In a smaller but non-negligible set of cases, the model misinterprets the intent of the question and produces an answer to a different problem than the one asked. This often appears as computing the wrong quantity (e.g., total interest instead of total amount owed). %These errors reflect failures in task understanding rather than numerical reasoning, and are most common in questions involving APRs and fees, where similar terms and subtle wording differences can cause the model to solve a different problem.

\section{What Makes Financial Literacy Questions Difficult? (RQ3)}
\label{sec:lr}
While question difficulty is often equated with the number of reasoning steps, financial literacy errors frequently arise from misapplied rules, overlooked conditions, and linguistic ambiguity. We quantify these effects using a regression framework over GPT-OSS-120B responses that isolates reasoning, semantic, and agreement-level factors to better understand what makes a question harder to solve.

\subsection{Model Specification}

We model the probability that an LLM answers a question correctly using a logistic regression, where $y_i \in \{0,1\}$ indicates whether the LLM answers question $i$ correctly,
where $y_i = 1$ denotes a correct answer. The model is specified as:
\begin{equation}
\log \frac{\Pr(y_i = 1)}{1 - \Pr(y_i = 1)}
=
\boldsymbol{\beta}^{\top} \mathbf{x}_i
+
\boldsymbol{\delta}^{\top} \mathbf{z}_i
+
\theta\,\texttt{third\_person}_i
+
\gamma_{c(i)} + \epsilon_i
\label{eq:model_spec}
\end{equation}

\textbf{Structural Reasoning Features: \textit{How does the structure of the solution program affect model performance?}} In Equation~\ref{eq:model_spec}, the vector $\mathbf{x}_i$ captures structural properties of the reasoning required to solve question $i$, as defined by its annotated solution program. We distinguish between \textit{global} and \textit{local} structural properties.Global features summarize aggregate characteristics of the solution program, including the total number of steps and the number of unique operators used. Meanwhile, local features capture the presence of specific operators, including indicators for conditional branching and comparison operations, as well as the use of key operators such as division, max/min, rounding, and exponentiation.

\textbf{Linguistic Framing: \textit{Does the way a question is asked matter?}} The role of question phrasing remains relatively underexplored in both financial and mathematical reasoning tasks. Through \texttt{third\_person} we focus on a specific dimension of linguistic framing: whether a question is written in the third person as opposed to the first person. This distinction may reflect the difference between a more realistic, personalized query, where an individual asks about their own financial situation, and a more detached, textbook-style formulation, which is more commonly used in existing datasets.

\textbf{Input and Context: \textit{How do input unit types and question context affect model performance?}} The vector $\mathbf{z}_i$ captures input-related and contextual characteristics of question $i$. We distinguish between \textit{input types} and \textit{contextual features}. Input unit types include binary indicators for the presence of different numerical inputs in the question. Specifically, we consider monetary values (e.g., dollar amounts), percentage-based quantities, and temporal inputs such as days, months, or years. Meanwhile, contextual features capture aspects of the financial setting described in the question. These include the number of distinct financial terms referenced, as well as indicators for specific contextual conditions such as hardship-related clauses.

\textbf{Credit-card agreement effects:}
The term $\gamma_{c(i)}$ represents agreement-specific adjustments, where $c(i)$ indexes the credit-card agreement associated with question $i$. Intuitively, this allows each agreement to have its own baseline difficulty, capturing differences such as fee structures, wording, and document organization. This allows the model to capture the inherent difficulty of each agreement separately, ensuring that the effects of question features are not driven by differences in agreement structure.

\subsection{Regression Results}
The full set of coefficient estimates and associated p-values is reported in Table~\ref{tab:regression_full} of Appendix~\ref{ap:logistic_regression}. Here, we discuss the largest and most significant effects.

\textbf{Reasoning structure and comparison operations.}
The number of required computational steps is positively associated with correctness ($\beta=0.33$, $p=0.012$), corresponding to approximately a 39\% increase in the odds of a correct answer per additional step. This does not suggest that longer questions are inherently easier, but rather that multi-step structure often forces explicit application of contractual conditions. This aligns with our error analysis, which shows that models frequently fail due to overgeneralization of default rules.

In contrast, comparison-based reasoning substantially hurts performance ($\beta=-1.33$, $p<0.001$) the largest among all features, reducing the odds of a correct answer by approximately 74\%. Notably, holding other factors constant, removing comparison operations would increase the odds of correctness by nearly 4 times. Such questions require selecting among competing fees or rates, where exceptions determine the outcome. These comparison-heavy scenarios are common in penalty- or restriction-driven settings and therefore represent a common source of failure, particularly for users in financially adverse contexts.

\textbf{Using First-Person Framing.}
Linguistic framing plays a meaningful role in model accuracy. We find that first-person phrasing is associated with significantly higher performance, while third-person prompts ($\beta=-0.55$, $p=0.005$) correspond to an approximate 42\% reduction in the odds of a correct response. One plausible explanation is that first-person framing provides a stronger signal that the query constitutes a direct user instruction rather than a descriptive or narrative task. Because instruction-tuned datasets frequently contain first-person queries (e.g., “How much will I owe if...?”), this framing more closely matches the distribution on which models are trained. Recent work on alignment faking further demonstrates that LLMs adjust their behavior in response to subtle contextual cues about training or deployment settings \citep{greenblatt2024alignmentfakinglargelanguage,needham2025largelanguagemodelsknow}. 

\textbf{Monetary values, conditional logic, and agreement complexity.}
Questions involving explicit monetary quantities are significantly less likely to be answered correctly ($\beta=-0.50$, $p=0.004$), even after controlling for reasoning structure and numerical operations. This suggests that difficulty arises not from arithmetic alone, but from correctly interpreting how monetary values interact with contractual rules such as minimum charges, caps, or penalty triggers. Conditional branching further reduces performance ($\beta=-0.59$, $p=0.046$), reflecting challenges with decision-dependent computation and multi-path logic.

These features frequently co-occur in scenarios involving fees, penalties, and account restrictions, which concentrate more contractual conditions and exceptions and are more common in near-prime, secured, and private-label credit cards.

\textbf{Heterogeneity Across Credit Card Agreements}
After accounting for reasoning structure and semantic features through $\boldsymbol{\beta}$ and $\boldsymbol{\delta}$, we observe substantial variation in the agreement-level effects $\gamma_{c(i)}$, all of which are negative. We most often see statistically meaningful differences for cards designed for limited credit, store-branded retail cards, and rewards or brand-affiliated cards, with clear clustering based on card type. This suggests that financial literacy reasoning depends critically on agreement-specific rules and structure rather than general financial knowledge alone.
\section{Discussion}
Our results reveal critical reasoning gaps in how LLMs process the complex conditional logic inherent in credit card agreements, a finding that generalizes to broader contractual documents. While models can identify basic terms, their performance degrades significantly when faced with multi-step computations and contractual exceptions involving nested conditional operators. These limitations suggest that the ``democratization'' of financial well-being via AI is currently hindered by a brittleness in reasoning that carries heightened consequences for vulnerable consumers. 

\paragraph{Consumers, Developers, and Regulators: } Individual users, especially those in financially vulnerable positions, face the highest risk from model failures in edge-case scenarios like late-payment penalties. To address these gaps, LLM developers should prioritize symbolic reasoning frameworks such as Program-of-Thought (PoT), which we find narrows the logic-processing bottleneck. Furthermore, policymakers should consider standards for ``LLM-accessible'' disclosures; making credit card terms more easily interpretable by language models would ensure that automated tools provide consumers with more accurate and transparent financial guidance. 

\paragraph{Broader Impacts on Reasoning Research: } Beyond the financial domain, the performance discrepancy we observe between first-person and third-person query variants provides significant insights for broader language modeling research. Interestingly, our finding that first-person phrasing improves performance suggests that LLMs may exhibit a form of ``conversational advantage''. Given that models are heavily optimized through RLHF for assistant-like interactions, they may perform more effectively when queries are framed as natural, first-person requests rather than formal, third-person evaluation style. This underscores the importance of evaluating LLM reasoning through the lens of linguistic framing, ensuring that benchmarks accurately reflect model capabilities across diverse natural human-AI interaction styles.

\section{Related Work}

Recent advances in large language models (LLMs) have enabled strong performance on knowledge- and reasoning-intensive tasks \citep{huang-chang-2023-towards}, including coding \citep{jain2024livecodebench, nguyen2025codemmlu, loughridge2024dafnybench}, mathematics \citep{gulati2024putnamaxiom, fan2024hardmath, qiao-etal-2025-math, yang2025position}, and logic \citep{qi2025large, han-etal-2024-folio}. While existing math reasoning benchmarks report impressive results, they also reveal key limitations: performance is closely tied to training distributions and often fails to generalize across reasoning settings \citep{hong2025benchmarkingllmsmathematicalreasoning}. Moreover, models frequently exhibit memorization-like behavior, reusing solution strategies or outcomes from unmodified problems despite altered assumptions, suggesting that current benchmarks may overstate true reasoning ability \citep{huang2025mathperturb}. Additional related work is discussed in Appendix~\ref{sec:add_related_work}.

% The advancements of LLMs have unlocked significant opportunities for tackling knowledge and reasoning-intensive tasks \citep{huang-chang-2023-towards}. Such tasks include coding \citep{jain2024livecodebench, nguyen2025codemmlu, loughridge2024dafnybench}, mathematics \citep{gulati2024putnamaxiom, fan2024hardmath, qiao-etal-2025-math, yang2025position}, logic \citep{qi2025large, han-etal-2024-folio} and beyond. Existing math reasoning benchmarks have shown that LLMs can achieve impressive results, but they also highlight important limitations. Studies suggest that model performance in mathematical reasoning is often tied to the kinds of problems and data distributions the models were trained on, which means their abilities do not always generalize well across different types of reasoning tasks \citep{hong2025benchmarkingllmsmathematicalreasoning}.
% Moreover, models exhibit memorization-like behavior, ignoring modified assumptions and applying techniques from the original problem, or even reproducing the outcome of the unmodified version. These raise concerns that current benchmarks may overstate true reasoning ability \citep{huang2025mathperturb}. Other related works are discussed in Appendix \ref{sec:add_related_work}.
\section{Conclusion}

We introduce \textsc{CreditQA}, a benchmark for evaluating numerical reasoning of financial literacy, using real-world credit card agreements. These documents reflect the conditional, exception-heavy rules that govern everyday financial decisions. Our extensive evaluation across open- and closed-source models shows that methods such as PoT improve accuracy, but substantial challenges remain. Through error analysis and regression-based difficulty modeling, we show that model failures stem less from computation than from misapplied contractual rules, missed conditions, and document-specific complexity. These errors concentrate in fee-, penalty-, and restriction-heavy scenarios that disproportionately affect financially vulnerable users, underscoring the need for more robust, domain-aware reasoning in high-stakes financial literacy tasks.

\section{Acknowledgment}

We thank all annotators who contributed to the initial development of the questions and annotation guidelines. We also gratefully acknowledge the Partnership for an Advanced Computing Environment (PACE) at Georgia Tech, Snowflake, and Together AI for providing compute resources.

\section*{Ethics Statement}
Our work adheres to ethical considerations, although we acknowledge certain biases and
limitations in our study. Our artifacts are available through HuggingFace, and
GitHub under the CC-BY-NC-SA 4.0 license.

% \section*{Author Contributions}

% \section*{Acknowledgments}
% Use unnumbered first level headings for the acknowledgments. All

\bibliography{colm2026_conference}
\bibliographystyle{colm2026_conference}

\appendix

\section{Additional Related Works on LLM Reasoning}
\label{sec:add_related_work}

\paragraph{Improving LLM Reasoning}

Among the varying methodologies to improve LLM reasoning without modifying model parameters, many recent works have focused on restructuring inference itself. \citet{Imani2023MathPrompterMR} introduces a prompt-level structure to guide step-by-step mathematical reasoning and demonstrates significant improvements in performance compared to free-form reasoning. Moreover, \citet{Liang2024ImprovingLR} formalizes structured reasoning as an explicit exploration process by showcasing that reasoning accuracy scales  with inference compute. Subsequent work proposes an population-level evolution to favor stronger reasoning paths \citep{zhang2025populationevolveparallelsamplingevolutionary} and context pruning to create shorter, more focuses contexts \citep{Huang2024FewerIM} to boost model reasoning performance. 

In contrast to inference-time approaches, training-time reasoning induction directly modifies the model during training for stronger reasoning capabilities. \citet{Liu2025AgenticMathEL} improves mathematical reasoning by restructuring training data through agentic decomposition into sub-tasks. \citet{Wu2025Confucius3MathAL} applies reinforcement learning (RL) post-training and improves optimization objectives to create a lightweight model with competitive performances. Other works focus on refining internal evaluation signals: \citet{tian2025rectifyevaluationpreferenceimproving} aligns self-critique behavior with perplexity-aware rewards while \citet{Liu2025BeyondDP} extends a minimalist rule-based RL for reasoning to emerge without explicit CoT. These approaches showcase how data structuring and RL-based approaches to feedback signal and optimization objectives design can enhance reasoning performances.

\citet{Pan2023LogicLMEL} and \citet{Creswell2022SelectionInferenceEL} integrates deterministic symbolic solvers and selection-inference mechanisms to improve multi-step logical reasoning and interpretability. \citet{sun2024determlraugmentingllmbasedlogical} further refines premise selection by introducing three key modules to guide inferences from indeterminacy to determinacy. \citet{wang2025adaptiveselectionsymboliclanguages} enables dynamic selection for the most suitable symbolic language and further collaboration with external logic solvers. \citet{Miner2024ScheherazadeEC} investigates the consistency of CoT reasoning traces by logically chaining multiple math problems together. While these methods were used to help constrain reasoning, LLMs have also shown considerable performance improvements to question answer accuracy using structured knowledge graphs \citep{Yasunaga2021QAGNNRW}. Particularly, explicitly modeling reasoning as knowledge graphs have allowed for LLMs to enable multi-hop relational inference \citep{Choudhary2023ComplexLR} which can be further enhanced with non-linear reasoning graph representations \citep{Xu2024CRPRAGAR}. \citet{Wang2024JMLRJM} jointly trains retrieval and reasoning to improve medical-grounded question answering. \citet{Zhang2024LogiCodeAL} utilizes anomaly detection to validate reasoning outputs with rule-based explanations. 

Recent works in execution-augmented reasoning include the usage of Prolog programs and interpreters for exact answers \citep{Yang2024ArithmeticRW},  table-centric tools to aid with tabular reasoning tasks \citep{Lu2024TARTAO}, and lean code conversion for mathematical theorem proving \citep{breen2025axproverdeepreasoningagentic}. In application-driven settings, \citet{Wang2025FromST} combines Python code execution with formula retrieval to correct arithmetic errors and hallucinated formulas, while \citet{Deng2024OedipusLR} defines a domain-specific language to decompose CAPTCHA challenges into sub-tasks. \citet{Xu2024CRPRAGAR} combines reasoning graphs with retrieval and evaluation for complex logical queries. By relying on specialized tools and programs, LLMs can focus on reasoning and planning while offloading computation to reduce execution errors.

\paragraph{Evaluating Reasoning Behavior of LLMs}
While assessing final answer correctness is pivotal to any study, several works focus on evaluating the LLM reasoning processes to correct imbalanced evaluation preferences \citep{tian2025rectifyevaluationpreferenceimproving} or ensure consistency of CoT reasoning traces for intermediate steps \citep{Miner2024ScheherazadeEC}. Notably, \citet{Deb2023FillIT} reformulates math word problems into backward reasoning tasks to reveal significant drops in performances with inverse inference before proposing an ensemble approach. \citet{Macina2023MathDialAD} expands upon sustaining logical consistency by exploring the notion of a tutor-student math dialogue to capture reasoning behavior and interactions.

Several benchmarks and diagnostic tasks have been created to probe current LLMs capabilities in numerical and mathematical reasoning. \citet{Hwang2024EnhancingNR} introduces an operator-based reasoning process to analyze numeric manipulation errors. \citet{Li2024TargetedTF} studies how refinement of difficult numerical examples can improve benchmark performance. \citet{Rahman2025LargeLM} and its follow up work by \citet{Rahman2025AFN} tests basic arithmetic, advanced calculations, prime checking, and game of 24 to reveal that models fail systematically on tasks requiring heuristic search. Benchmarks have now been extended to evaluate numerical reasoning and real-world perceptual geometry in numeric computation \citep{zeng2025numinanaturalunderstandingbenchmark}.

\paragraph{Knowledge-Intensive Reasoning Tasks}
Transitioning LLM reasoning frameworks to the real-world domains, factors such as correctness, interpretability, and safety are emphasized. Recent works have assessed the roles LLMs could play within medical contexts, where \citet{Wang2025FromST} and \citet{Wang2024JMLRJM} improve medical QA by proposing MedRaC as an agentic system for evidence-based medical calculations and jointly training retrieval and reasoning to reduce calculation hallucination. \citet{Liu2025BeyondDP} shows that structured minimalist rule-based RL outperforms standard distillation and fine-tuning based approaches for complex medical questions. Similarly, \citet{Li2025BeyondCQ} creates a modular benchmark for chemical reasoning by formalizing chemistry tasks as multi-step chemical operations. Moreover, \citet{sanchez2025multillmcollaborationmedicationrecommendation} leverages LLMChemistry's concept to form ensembles of LLMs to generate, review, and evaluate medication recommendations.
Beyond medical contexts, recent work has expanded into geological exploration and mineral mapping \citep{yu-etal-2025-sta} and security-based applications \citep{Deng2024OedipusLR}.

\section{Additional Related Works on LLM Reasoning}
\label{ap:question_creation}

The following describes the steps taken during the question generation process. Annotators worked in small collaborative groups of three to four during question generation. Each group was responsible for approximately one unique credit card agreement per member. 

\textbf{Familiarization with Credit Card Agreements} To first understand the credit card agreement and its unique characteristics, annotators were required to read each credit card agreement in full and develop an understanding of the card's issuer, intended purpose, benefits, and target customer profile.  This was done to encourage annotators to consider how each card is intended to be used, resulting in more card-specific, realistic questions and greater diversity in the evaluation. We provide more details for this step in Appendix~\ref{ap:ap-familiarizing-cca}. 

\textbf{Identifying Financially Relevant Terms}
Once familiarized with the agreements, annotators identified financially relevant numeric terms appearing in each credit card agreement. Following the methodology of \citet{shah-etal-2024-numerical}, we identified financial numeric terms as any values expressed with a dollar sign (\$) or percentage sign (\%). This process also captures numeric values associated with time periods (e.g. billing cycles and grace periods), as well as monetary thresholds (e.g. credit limits and penalty thresholds), as these quantities directly determine when and how charges or percentage-based rates apply. Identifying these terms produced a structured knowledge base of financial concepts used to guide subsequent question generation. Annotators then researched each concept and wrote brief explanations, which were discussed within their groups and reviewed.

\textbf{Keeping Consistency with Prior Work}
We required all annotators to study two prior datasets: ConvFinQA~\cite{chen-etal-2022-convfinqa} and DOCMATH-EVAL~\cite{zhao-etal-2024-docmath}. These works provided concrete examples of well-structured financial numerical reasoning questions and operator-based solution formats. Annotators were permitted to begin creating questions only after demonstrating proficiency with the methods described in these works through two assignments, detailed in \autoref{ap:reading_assigment}. We use the operator-based domain-specific language (DSL) from ConvFinQA (proposed by \cite{chen-etal-2021-finqa}) to represent step-by-step solutions for \textsc{CreditQA}.

\textbf{Question Generation Process}
While annotators were encouraged to propose challenging and diverse reasoning questions, several requirements were imposed to ensure clarity and reliable evaluation. Each question was required to produce exactly one output, restricted to a single output type: currency, time, percentage, or categorical value. Furthermore, if any information required to answer a question is not present in the credit card agreement, the question must explicitly provide all necessary values and definitions. For example, if a question involves a currency exchange from currency A to currency B, the applicable exchange rate must be provided within the question. 

For each question, annotators were required to specify the financial terms involved, all input values and their corresponding units, which inputs were sourced from the credit card agreement (document inputs) versus provided by the user (user inputs), and a step-by-step solution expressed in the operator-based domain-specific language (DSL) consistent with FinQA.

\textbf{Final Quality Control and Curation.}
After annotator revisions, we performed a final review evaluating each question for correctness, clarity, grounding in the source agreement, and complete specification of terms, inputs, and operators. Questions failing these criteria were revised or removed.

\section{Familiarizing with Credit Card Agreements}
\label{ap:ap-familiarizing-cca}

For each credit card agreement, annotators were asked to understand the following three questions. 
\begin{itemize}
    \item \textbf{Card Identification.} What credit card is this agreement for? Describe the issuing bank and the associated company, noting that these may be distinct entities. Responses were expected to be informative and factually accurate.

    \item \textbf{Card Purpose and Benefits.} What is the intended purpose of this credit card? Describe any associated benefits (e.g., cash back, travel rewards, points programs) in detail, and reference where this information appears in the agreement.

    \item \textbf{Target Customer Profile.} Who is the ideal customer for this card? Discuss relevant preferences, needs, or lifestyle factors, as well as the likely socio-economic profile of potential cardholders (e.g., students, working professionals, higher-income consumers). Information about the card’s issuer and originator should inform this assessment.
\end{itemize}
After answering the questions, the annotators proceeded with identifying the numerical values involving a \textbf{\$} or \textbf{\%} sign and their associated terminology. All the unique terms and corresponding values for each agreement were then listed in a two-column table format with the term name on one side and the explanation of the term on the other. Through this step, differences in standard terms such as the APR ranges and inclusion of unique terms such as the Delta Skymiles' reward policy aided with diversifying questions to include card-specific policies. As the last step to prepare annotators for the CreditQA creation, annotators then utilized their understanding of numerical reasoning and card-specific terms to create five numerical reasoning questions per card agreement while following the operator notation methodology present in ConvFinQA. 

\section{ConvFinQA and DocMath-Eval Proficiency Assignment}
\label{ap:reading_assigment}
For each paper, annotators were required to conduct an ACL-type review structured with the following general questions to assist with providing a critical evaluation within their reviews.

\begin{itemize}
    \item \textbf{Paper Summary:} Summarize the paper by section, providing 2-3 lines for each section. The summaries were expected to be precise by providing key details such as model names, metrics, and references from the work itself.
    \item \textbf{Paper Strengths:} Identify the key strengths of the paper. Adequate evidence-based justifications were provided for each strength by referencing sections, figures, and tables from the paper.
    \item \textbf{Paper Weaknesses:} Discuss any weaknesses or limitations identified in the paper. The responses were likewise expected to provide evidence-based justifications from references to the paper.
    \item \textbf{Opportunities for Improvement:} Suggest potential improvements or future directions for the research. Annotators were instructed with finding related works and current literature to describe where the paper can be improved.
    \item \textbf{Potential Threats, Limitations, and Ethical Concerns:} Evaluate any potential threats to the validity, applicability, or ethical concerns surrounding the research. Discuss the limitations of the paper and address any ethical concerns that arise from the research or its potential applications.
\end{itemize}

To help formalize the paper's methodology in constructing numerical reasoning tasks, we extended the  \textbf{ConvFinQA review} to include the following questions.\\

\textbf{Task Formulation Questions:}
\begin{itemize}
    \item What were the main metrics used in the paper and what was the best performing model? Why do you think this model performed the best?
    \item What were the main models used and benchmarked against?
    \item What data was used in the study? Provide the data sources, format and other details about the data that was annotated.
    \item What are the inputs required to compute $P(A | T,B,Q_n)$, and how do previous questions $Q_0,Q_1,...,Q_{n-1}$ factor into this computation?
    \item What is the difference between Type I and Type II conversations?
\end{itemize}

\textbf{Testing Question Samples:}
\begin{itemize}
    \item  Use the "train.json" provided as apart of the dataset and choose two conversations to test using GPT: one Type I and one Type II. Ensure that the user prompts follow the paper's methodology to properly prompt the two conversation types. Then, take a screenshot of the respective conversation, including all the user prompts and generated responses. Report and comment on the performance of the model utilized.
\end{itemize}

\section{Example CreditQA Question}
\label{ap:question_format}

Table~\ref{tab:json_kv_cash_advance} shows a representative question from \textsc{CreditQA}.

% Preamble (add once)
% \usepackage{booktabs}
% \usepackage{array}
% \usepackage{fancyvrb}

\begin{table*}[t]
\centering
\small
\setlength{\tabcolsep}{8pt}

\begin{tabular}{p{2cm} p{11.5cm}}
\toprule
\textbf{Key} & \textbf{Value} \\
\midrule

\texttt{question} &
I am in urgent need of cash but get my paycheck in 15 days. How much can I withdraw
from my credit card at an ATM if I only want to use \$200 from my paycheck to pay back
what I borrowed? The bank cash advance APR is 29.24\%. \\

\texttt{ground\_truth} & \texttt{188} \\
\texttt{output\_units} & \texttt{\$} \\

\texttt{input\_units} &
\begin{Verbatim}[fontsize=\small]
{
  "from_doc":  { "%": 1, "month": 0, "day": 0, "year": 0,
                 "$": 0, "binary": 0, "categorical": 0 },
  "from_user": { "%": 1, "month": 0, "day": 1, "year": 0,
                 "$": 1, "binary": 0, "categorical": 0 },
  "total":     { "%": 2, "month": 0, "day": 1, "year": 0,
                 "$": 1, "binary": 0, "categorical": 0 }
}
\end{Verbatim}
\\

\texttt{steps} &
\begin{Verbatim}[fontsize=\small]
[
  { "op": "divide",   "arg_1": "5",     "arg_2": "100", "resp": 0.05 },
  { "op": "add",      "arg_1": "1",     "arg_2": "#0",  "resp": 1.05 },
  { "op": "divide",   "arg_1": "29.24", "arg_2": "365", "resp": 0.0801 },
  { "op": "multiply", "arg_1": "#2",    "arg_2": "15",  "resp": 1.2016 },
  { "op": "divide",   "arg_1": "#3",    "arg_2": "100", "resp": 0.0120},
  { "op": "add",      "arg_1": "1",     "arg_2": "#4",  "resp": 1.0120 },
  { "op": "multiply", "arg_1": "#1",    "arg_2": "#5",  "resp": 1.0626},
  { "op": "divide",   "arg_1": "200",   "arg_2": "#6",  "resp": 188.213 }
]
\end{Verbatim}
\\

\texttt{program} &
\begin{Verbatim}[fontsize=\small]
[
  "divide(5, 100)",
  "add(1, #0)",
  "divide(29.24, 365)",
  "multiply(#2, 15)",
  "divide(#3, 100)",
  "add(1, #4)",
  "multiply(#1, #5)",
  "divide(200, #6)"
]
\end{Verbatim}
\\

\texttt{question\_tense} & \texttt{FIRST} \\

\texttt{financial\_terms} &
\begin{Verbatim}[fontsize=\small]
["CASH ADVANCE APR"]
\end{Verbatim}
\\

\texttt{credit\_card} & \texttt{Barclaycard\_combined.md} \\

\bottomrule
\end{tabular}

\caption{Example \textsc{CreditQA} question rendered as a key–value table. In the dataset, this record is stored as a structured JSON object with nested fields. Numbers are rounded for formatted.}
\label{tab:json_kv_cash_advance}
\end{table*}

\hypersetup{
    colorlinks,
    linkcolor={red!50!black},
    citecolor={blue!50!black},
    urlcolor={blue!80!black}
}
\tcbset{
  systembox/.style={
    colback=blue!5!white, colframe=blue!75!black,
    boxrule=0.5pt, arc=2pt, left=6pt, right=6pt, top=6pt, bottom=6pt
  },
  userbox/.style={
    colback=gray!10!white, colframe=black!50,
    boxrule=0.5pt, arc=2pt, left=6pt, right=6pt, top=6pt, bottom=6pt
  }
}

\section{Experiment Settings}
\label{ap:experiment-settings}

\begin{figure*}
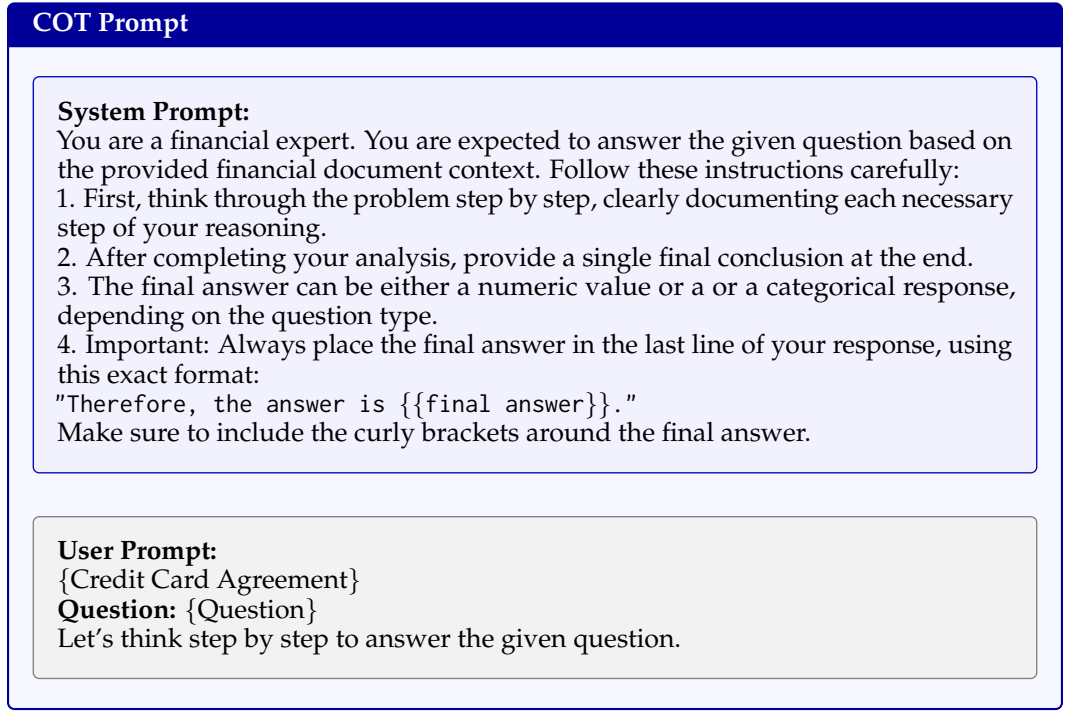

\centering

\begin{tcolorbox}[
  colback=blue!3!white,
  colframe=blue!60!black,
  title=\textbf{COT Prompt},
  boxrule=0.8pt,
  arc=2pt,
  left=6pt, right=6pt, top=8pt, bottom=8pt
]

\begin{tcolorbox}[systembox]
\textbf{System Prompt:} \\
You are a financial expert. You are expected to answer the given question based on the provided financial document context. Follow these instructions carefully:

1. First, think through the problem step by step, clearly documenting each necessary step of your reasoning. \\
2. After completing your analysis, provide a single final conclusion at the end. \\
3. The final answer can be either a numeric value or a or a categorical response, depending on the question type. \\
4. Important: Always place the final answer in the last line of your response, using this exact format: \\
\quad \texttt{"Therefore, the answer is \{\{final answer\}\}."} \\
Make sure to include the curly brackets around the final answer.
\end{tcolorbox}

\vspace{0.5em} % spacing between inner boxes

\begin{tcolorbox}[userbox]
\textbf{User Prompt:} \\
\{Credit Card Agreement\}

\textbf{Question:} \{Question\}

Let's think step by step to answer the given question.
\end{tcolorbox}

\end{tcolorbox}

\caption{Chain-of-Thought (COT) prompt template.}
\label{fig:cot_prompts}
\end{figure*}

\begin{figure*}[t]
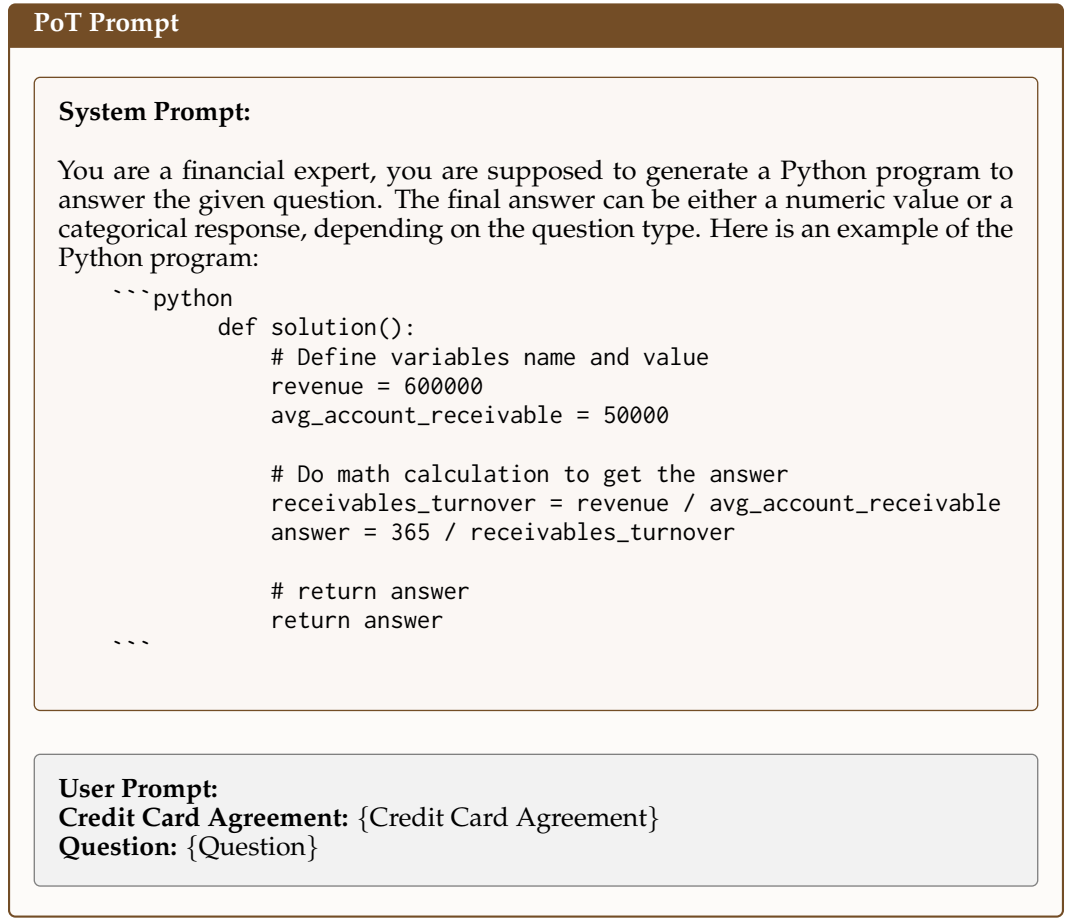

\centering

\begin{tcolorbox}[
  width=\textwidth,
  colback=brown!3!white,
  colframe=brown!60!black,
  title=\textbf{PoT Prompt},
  boxrule=0.8pt,
  arc=2pt,
  left=6pt,
  right=6pt,
  top=8pt,
  bottom=8pt
]

\begin{tcolorbox}[systembox, colback=brown!6!white, colframe=brown!60!black]
\textbf{System Prompt:} \\

You are a financial expert, you are supposed to generate a Python program to answer the given question. The final answer can be either a numeric value or a categorical response, depending on the question type. Here is an example of the Python program:
    \begin{verbatim}
    ```python
            def solution():
                # Define variables name and value
                revenue = 600000
                avg_account_receivable = 50000
                
                # Do math calculation to get the answer
                receivables_turnover = revenue / avg_account_receivable
                answer = 365 / receivables_turnover
                
                # return answer
                return answer
    ```
    \end{verbatim}
\end{tcolorbox}

\vspace{0.5em}

\begin{tcolorbox}[userbox]
\textbf{User Prompt:} \\
\textbf{Credit Card Agreement:} \{Credit Card Agreement\}

\textbf{Question:} \{Question\}
\end{tcolorbox}

\end{tcolorbox}

\caption{Program-of-Thought (PoT) prompt template.}
\label{fig:pot_prompt}
\end{figure*}

For all models, we set temperature to 0. This is consistent with prior literature \cite{tang-etal-2025-financereasoning, zhao-etal-2024-knowledgefmath}. Following \cite{tang-etal-2025-financereasoning}, we set the max tokens to 8192 as well. Figures~\ref{fig:cot_prompts} and~\ref{fig:pot_prompt} show the prompt template for CoT and PoT, respectively.

\section{Answer Extraction: Manual Annotation}
\label{ap:manual_annotation}

After extracting model-generated answers using regular-expression–based rules, we apply normalization and standardization procedures consistent with prior work \cite{zhao-etal-2024-knowledgefmath}. These procedures include numeric parsing, unit normalization, and tolerance-based matching. However, unlike many existing numerical reasoning benchmarks, \textsc{CreditQA} contains a non-trivial proportion of categorical answers (approximately 19.8\%), such as \emph{minimum vs.\ percentage-based fees} or \emph{policy-driven outcomes}. In such cases, purely numeric normalization is insufficient.

To address this, we apply additional string normalization heuristics (e.g., synonym collapsing and unit canonicalization). When automatic matching remains ambiguous—typically due to unit mismatches, categorical reasoning, or paraphrased but semantically equivalent outputs—we manually review the affected questions and assign correctness labels.

\paragraph{Coverage and Scope.}
Manual annotation was required for only a small subset of questions. Across all models, each LLM produced valid extracted answers for at least 93\% of the dataset. Importantly, unanswered or ambiguous cases were not consistent across models: questions unanswered by one model were often successfully answered by others. As a result, these cases do not systematically bias cross-model comparisons.

The number of manually annotated questions per model ranged from approximately 30 to 100 in most settings. The primary exception was Qwen under Program-of-Thought (PoT) prompting, which required manual annotation for slightly more than 150 questions. Even in this case, manual review affected a minority of the total evaluation set. We provide an example to help  clarify when and why manual annotation was applied:

\textit{``If I make a transaction of \$1000, will I pay \$5 or 5\% of my transaction in transaction fees?''}

For the Citi Strata Premier card, the agreement specifies a transaction fee of \emph{the greater of \$5 or 5\% of the transaction amount}. The ground-truth answer in our dataset is categorical (``5\%''). However, Qwen produced a numeric answer of ``\$50.'' Since 5\% of \$1000 equals \$50, which exceeds \$5, the model’s response is numerically consistent with the correct categorical rule. We therefore manually annotated this response as correct.

\paragraph{Annotation Criterion.}
Manual annotation was applied exclusively in cases where the model output and the ground-truth answer differed in representation—most commonly due to unit mismatches, categorical-versus-numeric expressions, or benign paraphrasing—rather than substantive disagreement. No manual corrections were made to override genuinely incorrect reasoning or calculations.

\section{Glossary of Financial Terms}
\label{ap:glossary}

This glossary defines key financial terms as they are generally used in credit card agreements and throughout our evaluation. Definitions reflect contractual usage rather than informal or colloquial interpretations.

\paragraph{Annual Percentage Rate (APR).}
The annualized interest rate applied to outstanding balances, expressed as a percentage. Credit card agreements may specify multiple APRs depending on transaction type and account status.

\paragraph{Purchase APR.}
The APR applied to standard purchase transactions. This rate typically differs from APRs applied to cash advances or penalty conditions and is often subject to a grace period.

\paragraph{Cash Advance APR.}
A distinct APR applied to cash advance transactions, typically higher than the Purchase APR and often accruing interest immediately without a grace period.

\paragraph{Penalty APR.}
A higher APR applied only after specific triggering events defined in the agreement, such as late payments beyond a grace period. Penalty APRs are conditional and should not be applied unless triggering criteria are met.

\paragraph{Grace Period.}
The period during which interest does not accrue on new purchases if the full statement balance is paid by the due date. Grace periods generally apply only to purchase balances and not to cash advances or penalty balances.

\paragraph{Billing Cycle.}
The fixed time interval over which transactions are aggregated to produce a statement balance, typically lasting 28--31 days.

\paragraph{Statement Balance.}
The total balance reported at the end of a billing cycle. Interest calculations, minimum payment requirements, and grace period eligibility are often defined relative to this balance.

\paragraph{Minimum Payment.}
The minimum amount a cardholder must pay by the due date to keep the account in good standing. This amount is often defined as the maximum of a fixed dollar minimum and a percentage of the outstanding balance plus fees and interest.

\paragraph{Minimum Fee.}
A fixed fee applied under certain conditions (e.g., late payment fees) that does not accrue interest unless explicitly stated in the agreement.

\paragraph{Interest Accrual.}
The process by which interest is added to an outstanding balance over time, either via simple or compound interest as specified in the agreement.

\paragraph{Simple Interest.}
Interest calculated solely on the principal balance, without compounding. This is uncommon for revolving credit balances unless explicitly stated.

\paragraph{Compound Interest.}
Interest calculated on both the principal and previously accrued interest. Credit card interest typically compounds daily unless otherwise specified.

\paragraph{Fixed Fee.}
A non-percentage-based charge (e.g., late payment fee or returned payment fee) that is applied as a flat amount rather than as interest on a balance.

\paragraph{Conditional Logic.}
Contractual rules that apply only when specific conditions are met (e.g., thresholds, timing constraints, eligibility criteria). Failure to correctly identify or apply these conditions leads to many observed model errors.

\paragraph{Transaction Category.}
The classification of a charge (e.g., purchase, cash advance, balance transfer) that determines which APRs, fees, and rules apply.

\section{Defining Difficulty In Financial Literacy Numerical Reasoning}
\label{ap:logistic_regression}

\begin{table*}[t]
\centering
\small
\setlength{\tabcolsep}{6pt}
\renewcommand{\arraystretch}{1.05}

\begin{tabular}{l c c}
\toprule
\textbf{Variable} & \textbf{Coefficient} & \textbf{\emph{p}-value} \\
\midrule
\multicolumn{3}{l}{\textit{Structural Reasoning Features}} \\
Total number of operator steps ($\texttt{num\_steps}$) & 0.332$^{**}$ & 0.012 \\
Unique operators ($\texttt{unique\_ops}$) & 0.040 & 0.710 \\
Conditional / branching logic ($\texttt{has\_branch}$) & $-$0.595$^{**}$ & 0.046 \\
Comparison operation ($\texttt{has\_compare}$) & $-$1.334$^{***}$ & $<\!0.001$ \\
Division operation ($\texttt{has\_divide}$) & 0.290 & 0.128 \\
Exponentiation ($\texttt{has\_exponent}$) & 0.263 & 0.363 \\
Rounding operation ($\texttt{has\_round}$) & $-$0.080 & 0.666 \\
Max / min operation ($\texttt{has\_max}$) & $-$0.005 & 0.987 \\
Number of references ($\texttt{num\_references}$) & $-$0.189$^{*}$ & 0.064 \\

\midrule
\multicolumn{3}{l}{\textit{Input and Context Features}} \\
Contains monetary value ($\texttt{has\_money}$) & $-$0.495$^{***}$ & 0.004 \\
Contains percentage ($\texttt{has\_percent}$) & 0.000 & 0.999 \\
Contains time reference ($\texttt{has\_time}$) & $-$0.031 & 0.833 \\
Financial hardship--related terms ($\texttt{has\_hardship\_related\_terms}$)
 & $-$0.067 & 0.679 \\

\midrule
\multicolumn{3}{l}{\textit{Linguistic Framing Features}} \\
Third-person phrasing ($\texttt{is\_third\_person}$) & $-$0.550$^{***}$ & 0.005 \\
Number of financial terms ($\texttt{num\_financial\_terms}$) & 0.001 & 0.995 \\

\midrule
Pseudo $R^2$ & \multicolumn{2}{c}{0.148} \\
\bottomrule
\end{tabular}

\caption{
Full logistic regression results predicting model correctness.
Positive coefficients indicate increased log-odds of a correct response.
The dependent variable equals 1 if the model answer is correct.
$^{***}p<0.01$, $^{**}p<0.05$, $^{*}p<0.10$.
}
\label{tab:regression_full}
\end{table*}

\paragraph{Interpreting Regression Coefficients.}
Table~\ref{tab:regression_full} reports coefficients from a logistic regression, which quantify changes in the \emph{log-odds} of a model answering a question correctly. Positive coefficients indicate features associated with higher correctness, while negative coefficients indicate reduced correctness, with the magnitude reflecting the strength of the association.

Importantly, these coefficients do not indicate that the probability of answering correctly increases or decreases by a fixed amount. Instead, they describe how the \emph{odds} of correctness change relative to a baseline, holding all other variables constant. This formulation allows us to identify which features systematically make questions easier or harder for models, independent of the underlying difficulty of the question. As a result, odds ratios enable meaningful comparisons across heterogeneous question types and isolate the dominant factors driving model success or failure.

For interpretability, coefficients can be exponentiated to obtain odds ratios. For example, the coefficient on \texttt{num\_steps} is 0.332 ($p=0.012$), corresponding to an odds ratio of $e^{0.332}\approx1.39$. This implies that each additional reasoning step increases the odds of a correct response by approximately 39\%, holding all other variables constant. In contrast, the large negative coefficient on comparison-based reasoning (\texttt{has\_compare}, $-1.334$, $p<0.001$) corresponds to an odds ratio of $e^{-1.334}\approx0.26$, indicating a roughly 74\% reduction in the odds of correctness when comparison operations are required.

\section{Evaluating Retrieval Augmented Generation}
\label{ap:rag}
To evaluate whether retrieval improves long-context financial reasoning, we implement a retrieval-augmented generation (RAG) pipeline over credit card agreements.

\subsection{Sampled Questions and Model}

For evaluation, we randomly sample five questions from each credit card agreement, resulting in a total of 140 questions. All experiments are conducted using the GPT-OSS-120B model.

\subsection{Document Chunking}

Each credit card agreement is provided as a markdown document, typically spanning 13{,}000--20{,}000 tokens. Prior to retrieval, agreements are segmented into overlapping text chunks. Chunking is performed in two stages. First, documents are split along markdown section headers to preserve contractual structure (e.g., \textit{Fees}, \textit{Interest Charges}, \textit{Arbitration}). Second, long sections are further divided into fixed-size chunks targeting approximately 900 tokens, with an overlap of 160 tokens between adjacent chunks. To avoid overly small or excessively large contexts, chunks are constrained to contain between 120 and 1{,}200 words. This procedure balances contextual completeness with retrieval efficiency while preserving local rule dependencies.

\subsection{Index Construction}

For each chunk, we construct both lexical and semantic indexes. A BM25 index is built over chunks on a per-document basis to support exact and near-exact term matching for legally precise language. In parallel, each chunk is embedded using the OpenAI \texttt{text-embedding-3-large} model. Embeddings are L2-normalized and indexed using FAISS for efficient cosine similarity search. All indexes are precomputed and reused across experiments.

\subsection{Retrieval Procedure}

At inference time, retrieval is performed independently for each question and constrained to the corresponding credit card agreement. Candidate chunks are retrieved using both lexical and dense retrieval. Specifically, the top 30 chunks are retrieved using BM25 and the top 30 chunks are retrieved using dense embedding similarity. These candidate sets are combined using rank-based score fusion, with a dense retrieval weight of 0.55. From the fused ranking, the top 8 chunks are selected and concatenated to form the context provided to the model. This hybrid retrieval strategy balances sensitivity to exact contractual language with semantic generalization.

\subsection{Prompt Construction and Inference}

The retrieved chunks replace the full agreement text in the prompt, ensuring that the model observes only a small, question-conditioned subset of the document. We use the same POT prompt as used in our main experiments.

\subsection{RAG Hyperparameters}

We use the following hyperparameters throughout our experiments: chunk size of 900 tokens with 160-token overlap; 30 BM25 candidates and 30 dense candidates per query; 8 final retrieved chunks per question; dense retrieval fusion weight of 0.55; and OpenAI \texttt{text-embedding-3-large} for dense indexing.

\subsection{Results}

\begin{table}[t]
\centering
\small
\resizebox{\columnwidth}{!}{%
\begin{tabular}{lcc}
\toprule
\textbf{Error Threshold} & \textbf{Non-RAG Accuracy (\%)} & \textbf{RAG Accuracy (\%)} \\
\midrule
0.2\%  & 75.5 & 72.9 \\
5\%    & 85.5 & 80.1 \\
\bottomrule
\end{tabular}
}
\caption{Accuracy comparison between non-retrieval (Non-RAG) and retrieval-augmented (RAG) settings under two numeric error thresholds.}
\label{tab:rag_results}
\end{table}

Table~\ref{tab:rag_results} reports model accuracy under non-retrieval (Non-RAG) and retrieval-augmented (RAG) settings using our $\pm .2\%$ and $\pm 5\%$thresholds. Across both thresholds, the RAG setting exhibits slightly lower accuracy than the non-RAG baseline.

Manual inspection reveals that the performance decrease in the RAG setting is primarily driven by missing or incomplete contextual information in the retrieved excerpts. In particular, some questions require the identification of specific conditions, exceptions, or cross-sectional dependencies that are not fully captured within the retrieved chunks. As a result, the model occasionally applies incomplete contractual rules or overlooks qualifying conditions necessary for correct computation.

These findings suggest that, while retrieval constrains the model to evidence-grounded context, the chunking and retrieval process can omit critical contractual details.
\end{document}